\documentclass{article}
\usepackage[utf8]{inputenc}
\usepackage{graphicx}
\usepackage{algorithm} 
\usepackage{algpseudocode} 
\usepackage{xcolor}
\usepackage[a4paper, total={7in, 10in}]{geometry}
\usepackage{comment}
\usepackage{caption}
\usepackage{subcaption}
\usepackage{hyperref}

\title{Exploring Semantic Perturbations on Grover \\ 
    \large CMSC 473/673: Capstone in Machine Learning}
\author{Ziqing Ji, Pranav Kulkarni, Marko Neskovic, Kevin Nolan, Yan Xu}
\date{21 December 2021}

\begin{document}

\maketitle
\begin{center}
\url{https://github.com/itspranavk/cmsc473fall21-grover}
\end{center}

\begin{abstract}
    With news and information being as easy to access as they currently are, it is more important than ever to ensure that people are not mislead by what they read. Recently, the rise of neural fake news (AI-generated fake news) and its demonstrated effectiveness at fooling humans has prompted the development of models to detect it. One such model is the Grover model, which can both detect neural fake news to prevent it, and generate it to demonstrate how a model could be misused to fool human readers. In this work we explore the Grover model's fake news detection capabilities by performing targeted attacks through perturbations on input news articles. Through this we test Grover's resilience to these adversarial attacks and expose some potential vulnerabilities which should be addressed in further iterations to ensure it can detect all types of fake news accurately.
\end{abstract}
\section{Introduction and Literature}
\subsection{The Rise of Machine-Generated Fake News}
While the term “fake news” has become politically charged in recent years, researchers of online misinformation typically define it as a form of propaganda that knowingly transmits false information to readers, viewers, or users \cite{Boberg2019}. This definition is rather broad, with many academics offering their own take. Alcott and Gentzkow describe fake news as “news articles that are intentionally and verifiably false, and could mislead readers” \cite{Alcott2017}. Wardle and Derakhshan posit seven categories of fake news, which together make a scale ranging from simple satire to fully fabricated content \cite{Wardle2018}. For the purposes of this project, we will define fake news as any text that is created with the intention of imparting false information onto the reader to achieve the goal of the disseminator. This includes, as Zellers et Al. describe, “targeted propaganda that closely mimics the style of real news” \cite{Zellers2019-arXiv}. Taken this way, fake news has been present in social and political discourse for thousands of years. In one of the earliest cited examples of fake news, Ramses the Great of Ancient Egypt spread propaganda portraying the Battle of Kadesh as a glowing victory for Egypt over the Hittites. The Treaty of Kadesh reveals that in fact, the battle ended in stalemate \cite{Weir2009}. In the modern era, fake news can pose great threats in democracy, politics, economy, and so on \cite{Faris2017,Alcott2017,Vosoughi2018}. Creating and consuming high quality misinformation is notably easier over the internet compared to print media \cite{Shu2017,Alcott2017}. The current deep learning methods in natural language processing \cite{Devlin2018,Radford2019} can automatically generate texts, and they can be applied to fake news generation, making it harder to tell whether a given article is authentic or fake. This change in ease has caused a dramatic increase in the proliferation of fabricated content using social media \cite{Lazer2018}.

\subsection{An Algorithmic Response}
Lazer et al. asserted in 2018 that the rise of fake news prompts “a new system of safeguards” \cite{Lazer2018}. Social media companies have since adopted several strategies to detect and remove fake news from their websites. For obvious reasons, none of these strategies have been made publicly available. However, a good body of research on fake news detection is accessible to the wider public. Shu et Al. describe fake news detection from a data mining perspective. They focus more on social engagement, click-through rate, and share counts rather than the actual content of social media posts to determine what is legitimate \cite{Shu2017}. Reis et Al. created a new system of features and measures for use in supervised learning pertaining to fake news detection \cite{Reis2019}. Zhou et al. lay out a multidisciplinary approach in which they “broadly adopt techniques in data mining, machine learning, natural language processing, information retrieval, and social search” \cite{Zhou2019}. It is clear that there are a plethora of existing techniques to detect and address fake news in social media, but it is uncertain how well these techniques perform on real-world data. Additionally, it is unclear what vulnerabilities they possess. Our proposed paper will examine types of vulnerabilities in Grover, a state-of-the-art framework for detecting fake news.

\subsection{GPT-2 and Grover}
In early 2019, AI research company Open AI released GPT-2, a “large transformer-based language model with 1.5 billion parameters” \cite{Radford2019}. GPT-2 is remarkably good at synthesising new text given a prefix. While it still underperforms human ability, the model has broken algorithmic records in accuracy and perplexity on several datasets. The incredible accuracy of GPT-2 has led to some concern over its ethical use. Floridi and Chiriatti argue that with such technology,  “Readers and consumers of texts will have to get used to not knowing whether the source is artificial or human” [18]. Zeller et Al, however, focus on GPT’s implications on fake news \cite{Zellers2019-arXiv}. They conclude that GPT-2 and related technologies pose a significant threat to the validity of online information. The assumption is that GPT is highly capable of creating texts that fool both humans and machines, allowing for an even greater proliferation of fake news on the internet. To counter this threat, Zeller and his team created a state-of-the-art tool named Grover.  
Grover is both a neural fake news generator and detector. It is built using the same architecture as GPT2 \cite{Radford2019}. It considers five metadata fields of an article: domain, date, authors, headline, and body. The design of the model allows for flexible decomposition of the joint distribution. During inference, the model uses the fields as context where each field contains its content and field-specific tokens. The tokens are concatenated and appended with the target field-specific token to generate the target field. During training, they separate metadata fields into two sets F1 and F2, where they minimize the cross entropy loss of predicting tokens in F1 followed by F2. As is shown in the experiments, Grover noticeably improves on perplexity when conditioned on metadata, and it is over 5 perplexity points lower than GPT2 models. Humans are Easily Fooled by Grover-written propaganda. Also, Grover performs best at detecting Grover’s fake news compared with GPT2, BERT, and FastText \cite{Joulin2017}. Simple techniques aren’t enough to consistently fool Grover. For example, rejection sampling produces only a temporary advantage to a hypothetical attacker, because once Grover is trained on additional generations from that attacker, its accuracy returns to levels we expect \cite{Zellers2019-medium}.

\subsection{Vulnerabilities in Fake News Detection}
Currently, there is no substantial research on the robustness of fake news detectors. To fill this gap, we would like to research the vulnerabilities of Grover, as it is the best-performing detector to date. Specifically, we will design attacks at the word, sentence, paragraph, and article (i.e discourse) level to analyze Grover’s performance in detecting fake news. Some attacks include replacing single words with synonyms, rearranging phrases within sentences, and breaking up the “narrative structure” of an article. We hope to determine how much and what parts of a text need to be changed before Grover will miscategorize it. In conducting this research, we will provide analysis that can be used to make Grover and similar systems more robust, with the ultimate goal of improving validity in online texts.

\section{Getting Started}
\subsection{Setting up Grover}

As part of Zeller's goal of ensuring transparency in neural fake news detectors and generators, the Grover source code is publicly available on GitHub here: \url{https://github.com/rowanz/grover}

However, setting up Grover was one of the hardest parts of this project. The version of Grover available on GitHub has numerous inconsistencies and bugs that make it hard to set up and generate useful results. Some of these issues include:

\begin{itemize}
    \item Inconsistencies between naming conventions. For example, large and medium models, article and text fields, etc.
    \item Given discriminator does not consider article content to classify.
    \item Multiple authors are not supported. Authors must be converted from list to string.
    \item Setting up models for generation and discrimination was poorly documented.
\end{itemize}

As part of our research project, we will include a fixed version of the $run\_discrimination.py$ file in our GitHub repository that address some of the above concerns, in addition to providing an in-depth explanation of how to set up Grover for discrimination.

\subsection{Selection of Grover Model}

There are three main types of models offered by Zeller's team: Base, Large/Medium, and Mega. Each model can do generation and discrimination (using checkpoints). For our purpose, we decided to use the Large model because it provided a balance between accuracy and computational requirements. More specifically, we used the model \texttt{generator=medium $\sim$ discriminator=grover $\sim$ discsize=medium $\sim$ dataset=p=0.96} located in Zeller's Google cloud bucket. This enabled us not only to get valid results with $top-p = 0.96$, but also enabled us to run the model locally on our computers. Eventually, for larger scale tests, we used Google cloud TPUs to speed up the process.

\subsection{Analysis of Large Grover Model}

\begin{itemize}
    \item Vocabulary size: 50720
    \item Hidden Layers: 24
    \item Neurons per hidden layer: 1024
    \item Activation: GELU
\end{itemize} 

\section{Uninformed Perturbations}

The goal of making uninformed perturbations is to explore Grover's sensitivity to changes on word, sentence, and discourse-level. By tweaking some aspects of an article either randomly or contextually, we can study how Grover reacts to these changes. We also ensured article metadata remains consistent with the results we generate. Based on our observations, metadata like title, publish date and time, authors, and url have a significant impact in Grover's confidence when classifying. 

\subsection{"Frankenstein" Articles}
\label{sentence_swap}

The idea of creating "Frankenstein" articles is to intentionally break the flow of the article through substitution. We, as humans, are incredible good at detecting nonsensical article flow, but we want to explore Grover's sensitivity to such changes. There are two ways we can go about doing this:

\begin{itemize}
    \item Blending human and machine-written articles
    \item Blending articles of the same classification
\end{itemize}

Both methods test Grover's sensitivity to sentence-level and article i.e discourse-level changes. We define a "base" article as our target. We will control how much of this is retained. To generate these "Frankenstein" articles, we first choose our base i.e target article and our source articles. The source articles may be human-written or machine-written, depending on the context. We use NLTK's tokenize to split all article content into sentences. We randomize the selection of indices in the target and the indices in the source sentences. The goal is to substitute source sentences into our base article.

\begin{center}
    \includegraphics[width=\textwidth]{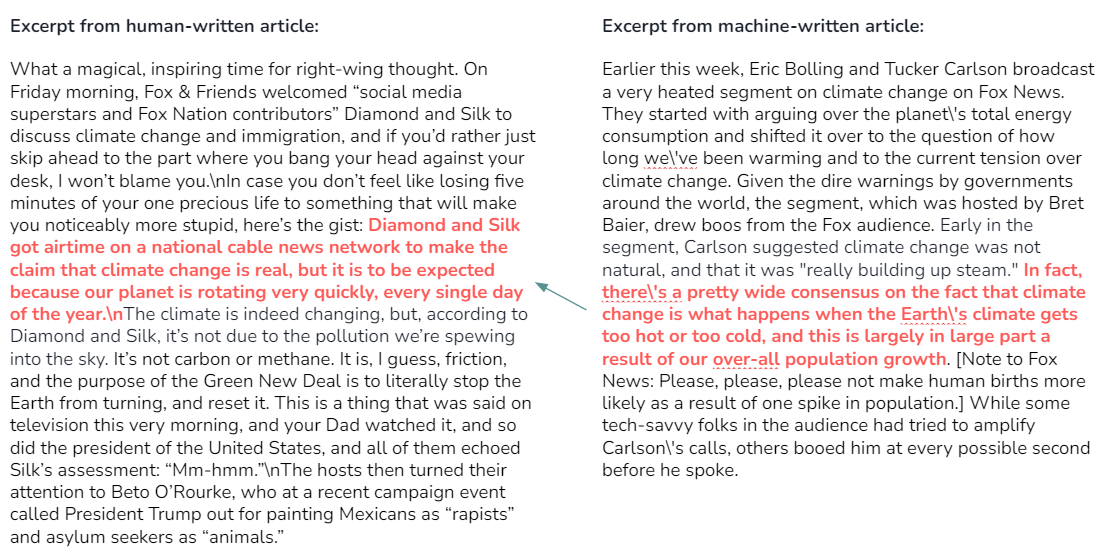}
\end{center}

\subsubsection{Blending Human-written and Machine-written Articles}

Blending human-written and machine-written articles allows us to control the percentage of fake content in an article and examine Grover's performance changes. The testing dataset we create in the end could be represented by a spectrum structure where the left-most side is the original human-written article and on the right-most side is the machine-written article that was generated. As we move along the spectrum form left to right, the proportion of fake content in the article increases, which allows us to see how Grover's confidence level in both class (Human vs. Machine) changes. 

\begin{figure}[htbp]
\centering
\includegraphics[width=15cm,height=10cm,keepaspectratio]{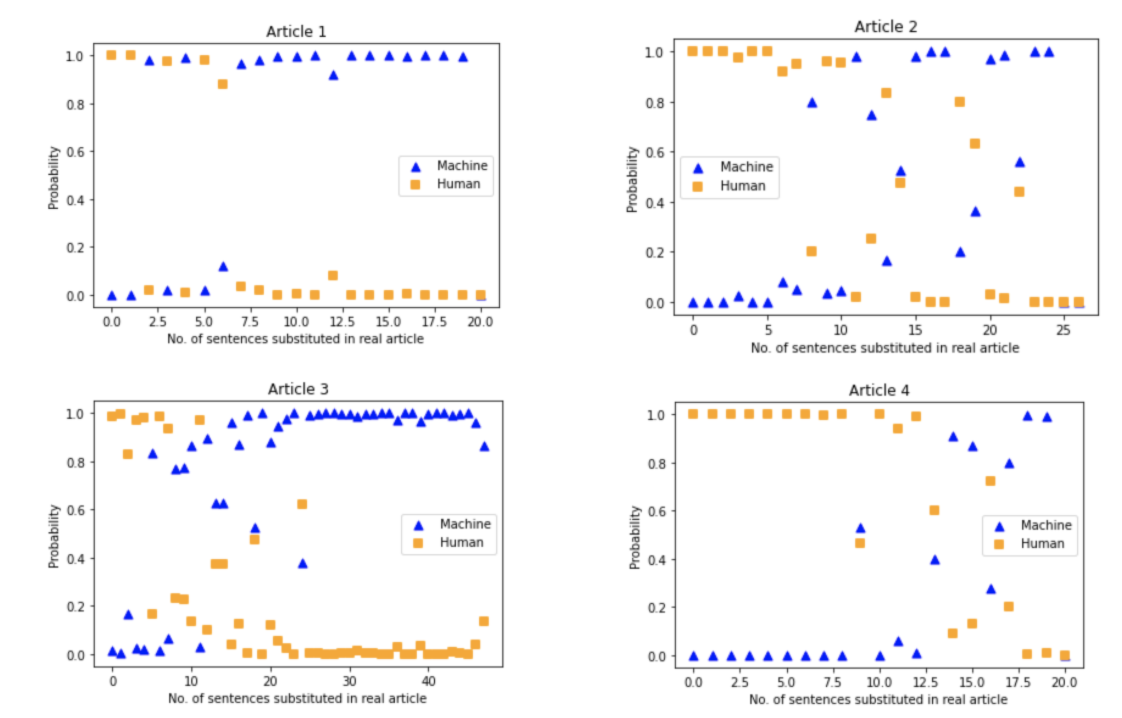}
\includegraphics[width=15cm,height=10cm,keepaspectratio]{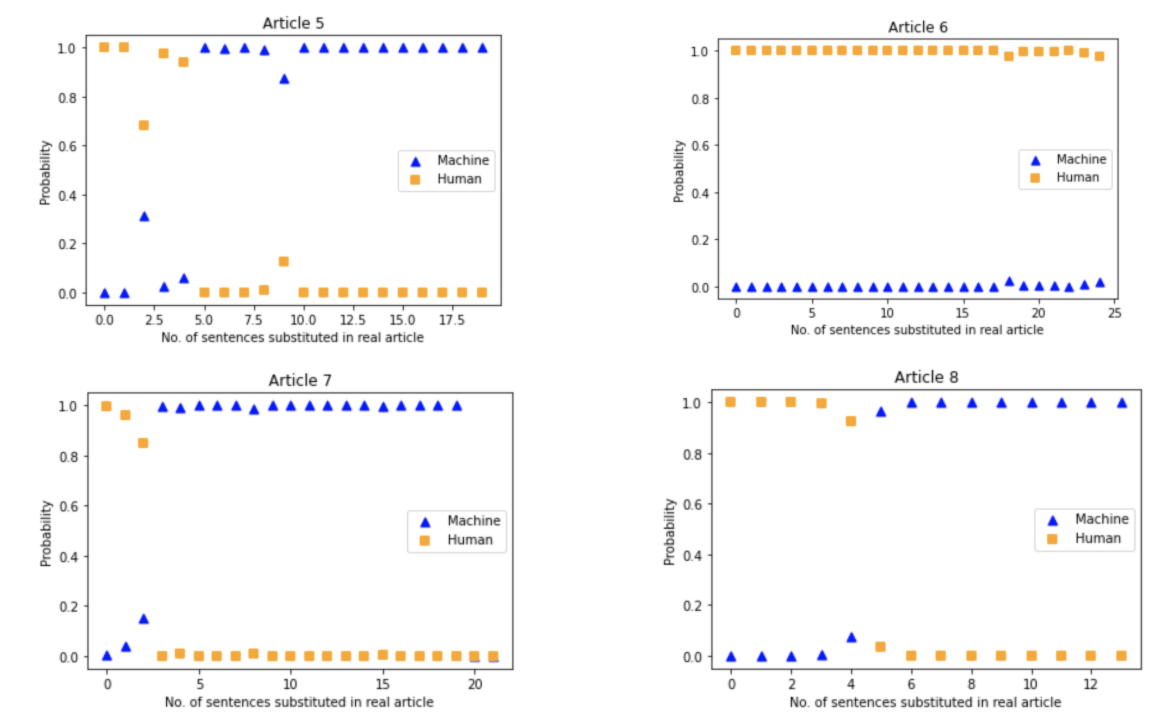}
\caption{Blending Human and Machine-written Articles, real-fake}
\label{fig:diffclass_real}
\end{figure}

According to the results in Figure \ref{fig:diffclass_real} we are able to see that as we move from left to right on the spectrum structure, there is a very clear boundary when Grover started to detect the article as a machine-written article, for example in articles 1, 7, and 8. However, there is not a very clear boundary in some other cases. For example, in articles 2, 3, and 4, there is a period of time where Grover hesitates, and makes back and forth decisions about whether to detected as human or machine. However, there is one exception case. In article 6, Grover detects all of versions of the articles as human, regardless of the percentage of fake sentences in the article, even the one that Grover itself generated.

\subsubsection{Blending Articles of the Same Classification}

\begin{figure}[htbp]
     \centering
     \begin{subfigure}[b]{0.4\textwidth}
         \centering
         \includegraphics[width=\textwidth]{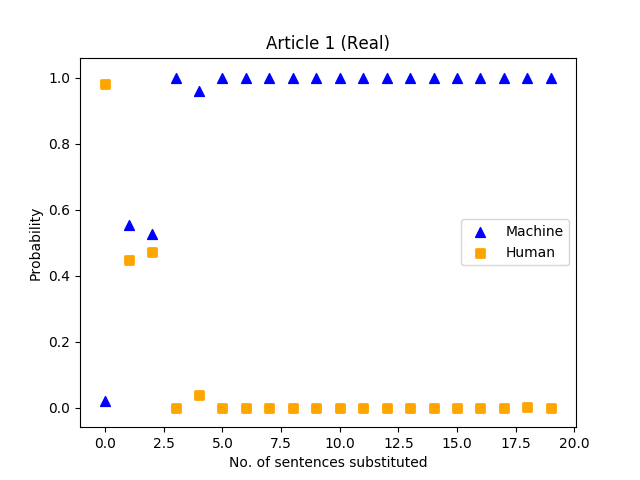}
         \caption{Article 1: Real}
     \end{subfigure}
     \hfill
     \begin{subfigure}[b]{0.4\textwidth}
         \centering
         \includegraphics[width=\textwidth]{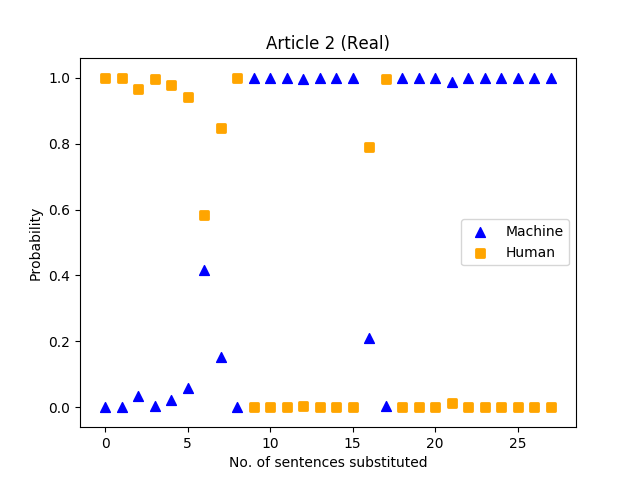}
         \caption{Article 2: Real}
     \end{subfigure}
     \begin{subfigure}[b]{0.4\textwidth}
         \centering
         \includegraphics[width=\textwidth]{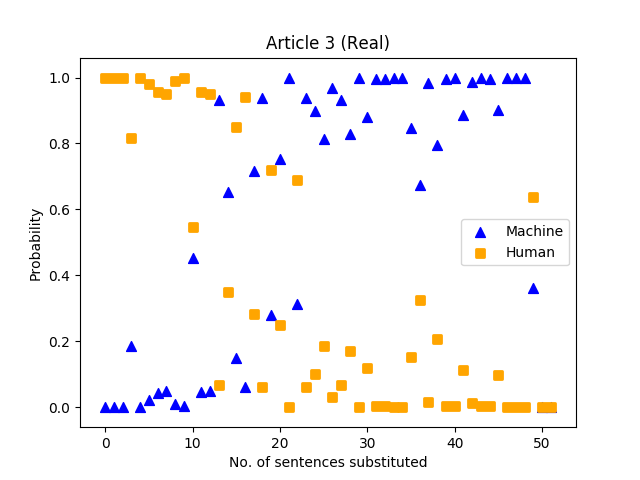}
         \caption{Article 3: Real}
     \end{subfigure}
     \hfill
    \begin{subfigure}[b]{0.4\textwidth}
         \centering
         \includegraphics[width=\textwidth]{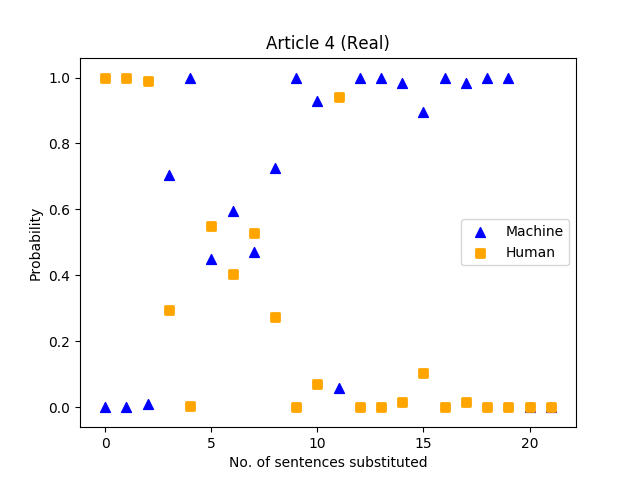}
         \caption{Article 4: Real}
     \end{subfigure}
    \begin{subfigure}[b]{0.4\textwidth}
         \centering
         \includegraphics[width=\textwidth]{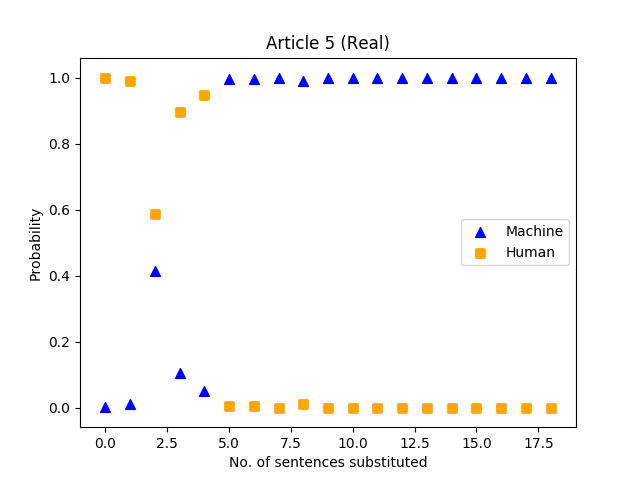}
         \caption{Article 5: Real}
     \end{subfigure}
     \hfill
    \begin{subfigure}[b]{0.4\textwidth}
        \centering
        \includegraphics[width=\textwidth]{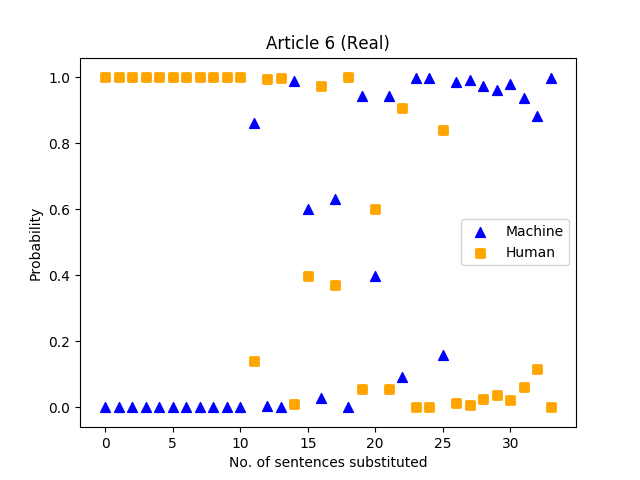}
        \caption{Article 6: Real}
    \end{subfigure}
    \begin{subfigure}[b]{0.4\textwidth}
         \centering
         \includegraphics[width=\textwidth]{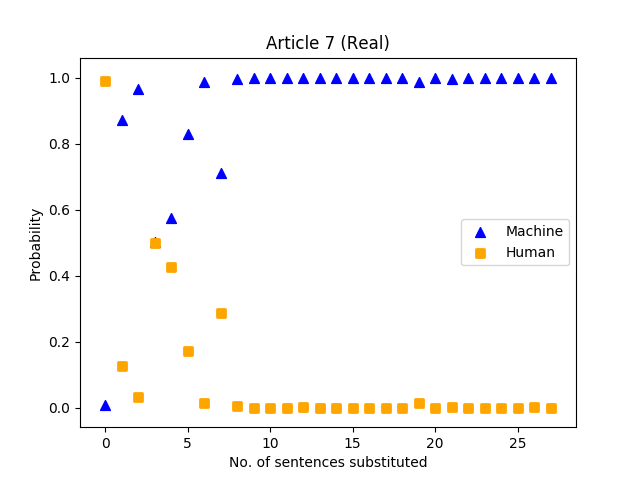}
         \caption{Article 7: Real}
    \end{subfigure}
    \hfill
    \begin{subfigure}[b]{0.4\textwidth}
         \centering
         \includegraphics[width=\textwidth]{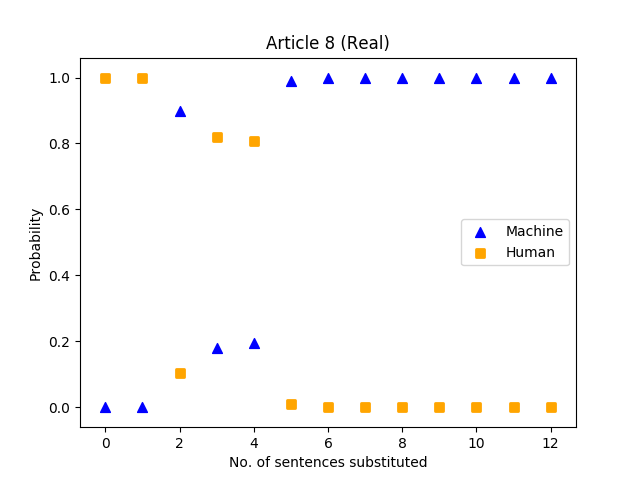}
         \caption{Article 8: Real}
     \end{subfigure}
    \caption{Blending articles with same classification, real-real}
    \label{fig:sameclass_real}
\end{figure}

Unlike the previous method, blending articles of the same classification aims to only test Grover's sensitivity to discourse-level changes. While we are still maintaining the context of the base article. However, an interesting fact here is that on the sentence-level, all generated "Frankenstein" articles are real. In the first part, we generated "Frankenstein" articles using purely real articles. After running the Grover discriminator, we got the following results in Figure \ref{fig:sameclass_real}.

Compared to the case where we blending human-written articles with machine-written ones, these results are more erratic in nature. In some cases, Grover easily recognizes the break in flow and can easily categorize the article as fake. We can see this happen with Articles 1, 5, 7, and 8. However, with Articles 2, 3 (especially here), 4, and 6, we can observe that in some cases Grover is unable to confidently classify as either human or machine-written. Grover flip-flops with both classes until eventually all articles are classified as fake. This is interested because Grover is not as confident in this situation as with the previous one.

\begin{figure}[htbp]
     \centering
     \begin{subfigure}[b]{0.4\textwidth}
         \centering
         \includegraphics[width=\textwidth]{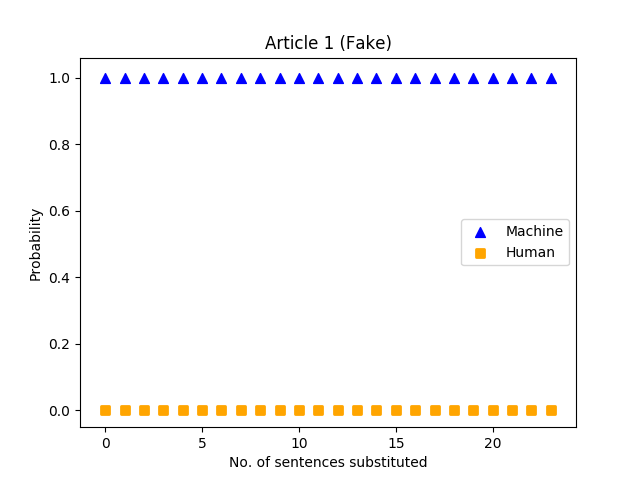}
         \caption{Article 1: Fake}
     \end{subfigure}
     \hfill
     \begin{subfigure}[b]{0.4\textwidth}
         \centering
         \includegraphics[width=\textwidth]{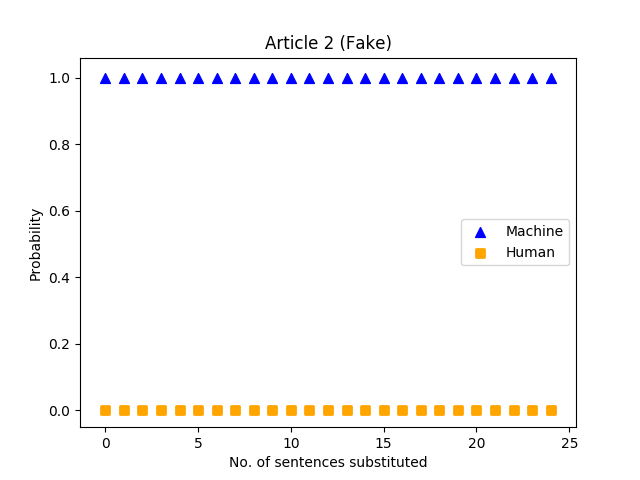}
         \caption{Article 2: Fake}
     \end{subfigure}
    \caption{Blending articles with same classification, fake-fake}
    \label{fig:sameclass_fake}
\end{figure}

Finally, we generated "Frankenstein" articles using purely fake articles. After running the Grover discriminator, we got the following results in Figure \ref{fig:sameclass_fake}.

We can clearly see that Grover is much more confident about classifying fake articles. At all points in the spectrum, Grover was confident about the "Frankenstein" articles being fake. A possible reason for this is that all articles are fake on sentence-level. In addition to discourse-level changes that break the flow of the article, we can say for sure that Grover is robust on both changes when the article is initially classified as fake.

\subsubsection{Comparing Insertion and Substitution}
\label{sentence_insertion}

\begin{figure}[htbp]
     \centering
     \begin{subfigure}[b]{0.4\textwidth}
         \centering
         \includegraphics[width=\textwidth]{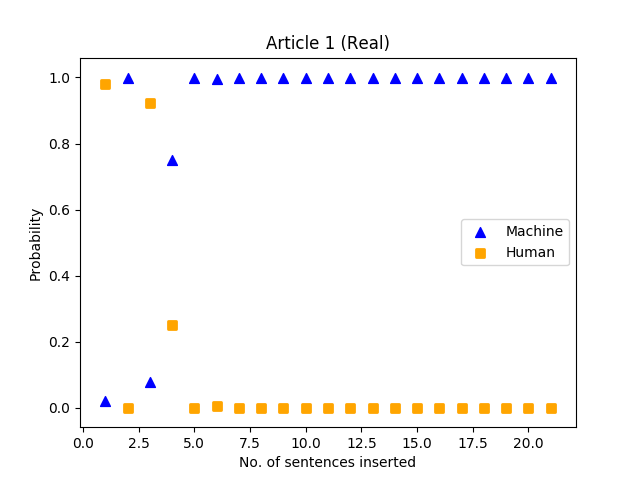}
         \caption{Article 1: Real}
     \end{subfigure}
     \hfill
     \begin{subfigure}[b]{0.4\textwidth}
         \centering
         \includegraphics[width=\textwidth]{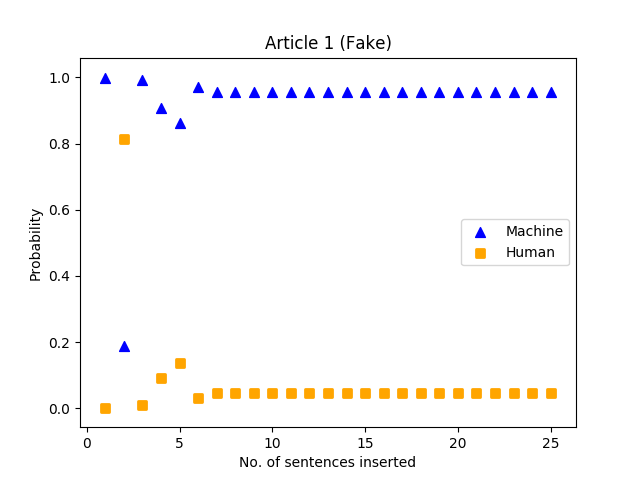}
         \caption{Article 1: Fake}
     \end{subfigure}
    \caption{Insertion instead of Substitution}
    \label{fig:insertion}
\end{figure}

In both previous methods, we focused on randomly substituting sentences in the base article to create human to machine spectrum. In this part, we explore whether insertion instead of substitution has an effect on Grover's confidence score. Here, we blended human and machine written articles. After running the Grover discriminator, we got the following results in Figure \ref{fig:insertion}.

Like we've seen previously with substitution, Grover is sensitive to discourse-level changes to real articles. We see a similar pattern where Grover can confidently classify our base article as fake after a certain number of changes. However, the way Grover reacts to insertion for fake articles is interesting. Previously, substituting sentences made no effect on Grover's confidence scores. With insertion, we can see that initially, Grover becomes "less sure" that the article was fake. In one case, Grover classified our fake article as real. Unfortunately, after a certain number of insertions, Grover can easily classify the article as fake.

\subsubsection{Effect of Position}

\begin{figure}[htbp]
\centering
\begin{subfigure}[b]{0.5\textwidth}
    \centering
    \includegraphics[width=\textwidth]{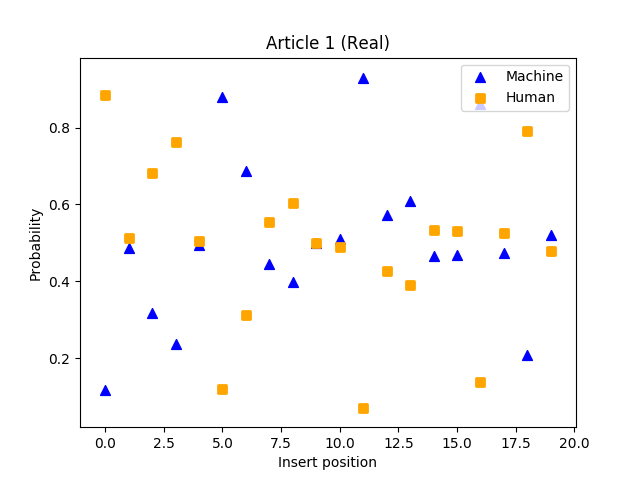}
    \end{subfigure}
    \caption{Effect of Position}
    \label{fig:position}
\end{figure}

Another aspect worthy of discussion is the role of position. In all previous tests, we randomized the position where sentences were substituted or inserted. In this part, we'll explore the effect of position on Grover's confidence score. We inserted a machine-written sentence in every possible position in a human-written article. We also ensured that the machine-written sentence was in the context of the article. After running the Grover discriminator, we got the results in Figure \ref{fig:position}

Overall in the results, we observed that position of insertion has an effect on the how Grover classifies an article. However, this effect has no clear pattern that could be exploited by adversaries. We can see that at some points, Grover is confident that the article is machine-written but looking at the actual article, this happens as a result of a break in the flow of the article at the start or end of each paragraph.

\subsubsection{Effect of Length}

\begin{figure}[htbp]
     \centering
     \begin{subfigure}[b]{0.4\textwidth}
         \centering
         \includegraphics[width=\textwidth]{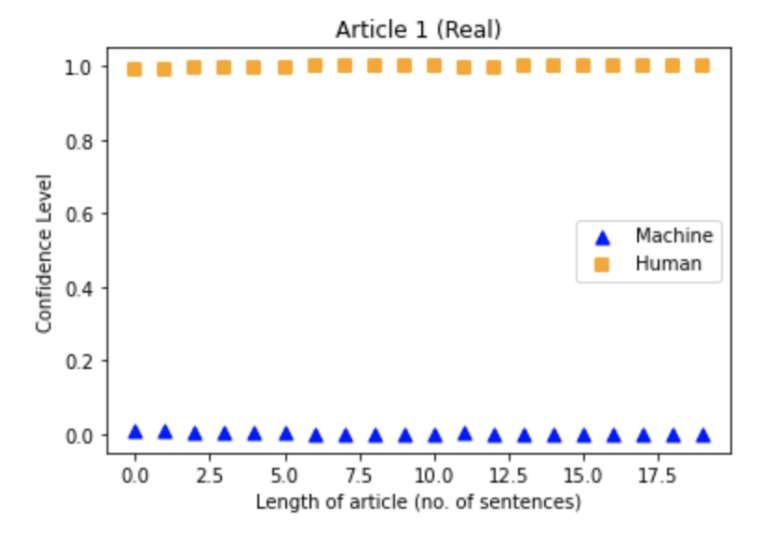}
         \caption{Article 1: Real}
     \end{subfigure}
     \hfill
     \begin{subfigure}[b]{0.4\textwidth}
         \centering
         \includegraphics[width=\textwidth]{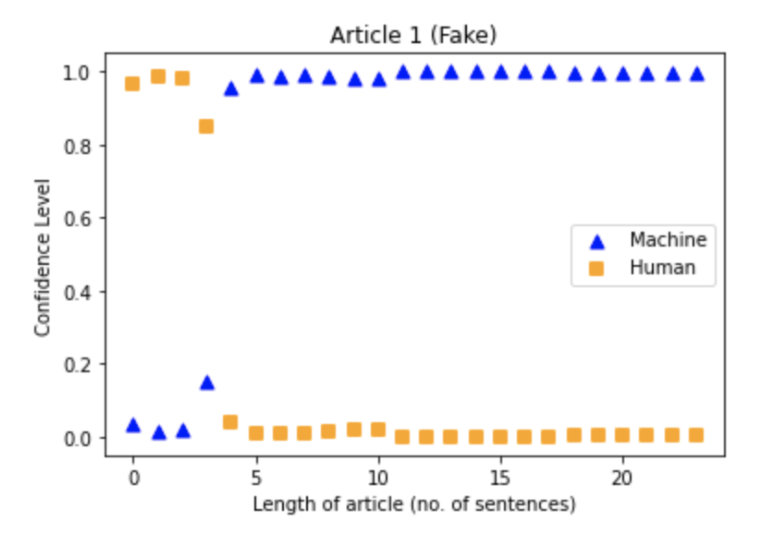}
         \caption{Article 1: Fake}
     \end{subfigure}
    \caption{Effect of Length}
    \label{fig:length}
\end{figure}

Whether different lengths of articles would lead to changes in Grover's performance was also examined. In this process, we did not involve any mixing of human-written and machine-written articles. Instead, we tested different lengths of the original human-written article and the original machine-written article. It is also worth to note that we kept the original order of the sentences in order to maintain the flow of the sentences. For example, the first sentence of the article is taken out and used as an input, and then the first two sentences, then the first three sentences etc. This would allow us to examine only the effect of length, while maintaining the other variables, such as flow of sentences etc. The results are shown in \ref{fig:length}.

Taking article 1 as an example, we can see that all the original human-written article with different lengths are classified as "human" as expected. However, some versions of the machine-written article was classified as "human". When the length of the machine-written article was less than 5 sentences, they were classified as "human" instead of "machine".

\subsubsection{Effect of Subjectivity}

We also wanted to examine the effect of subjectivity on the performance of Grover. The subjectivity here refers to whether a sentence is subjective or objective. Subjective sentences generally refer to personal opinions, emotions or judgements, whereas more objective sentences refer to facts. and we wanted to test whether inserting sentences of different subjectivity values have an effect on Grover's performance. 

\begin{figure}[htbp]
\centering
\begin{subfigure}[b]{0.5\textwidth}
    \centering
    \includegraphics[width=\textwidth]{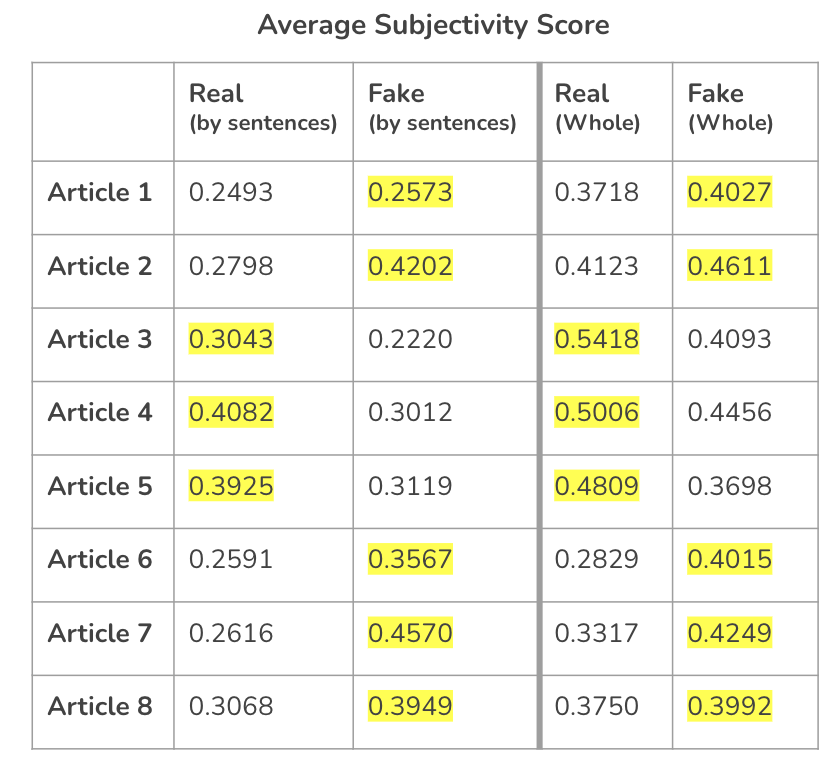}
    \end{subfigure}
    \caption{Overall Subjectivity}
    \label{fig:overall_subj}
\end{figure}

Before testing, we examined the overall subjectivity values of fake articles. The python package TextBlob was used to calculate the subjectivity value of the articles. Using this package, we would be able to receive a subjectivity value ranging [0:1]. The larger the value is, it means more subjectivity, and the smaller it is, it means less subjectivity. We used two different methods to calculate the overall subjectivity level. The first one is to get the subjectivity values for each sentence in the article and calculate the average of those values as the overall subjectivity value of the whole article (by sentences). The second way is to directly input the whole article to get the subjectivity value directly (whole). Out of the 8 articles from the mini testing dataset, 5 of them show that the generated machine-written article tend to be more subjective than the original human-written ones, as shown in 
\ref{fig:overall_subj}

\begin{figure}[htbp]
     \centering
     \begin{subfigure}[b]{0.4\textwidth}
         \centering
         \includegraphics[width=\textwidth]{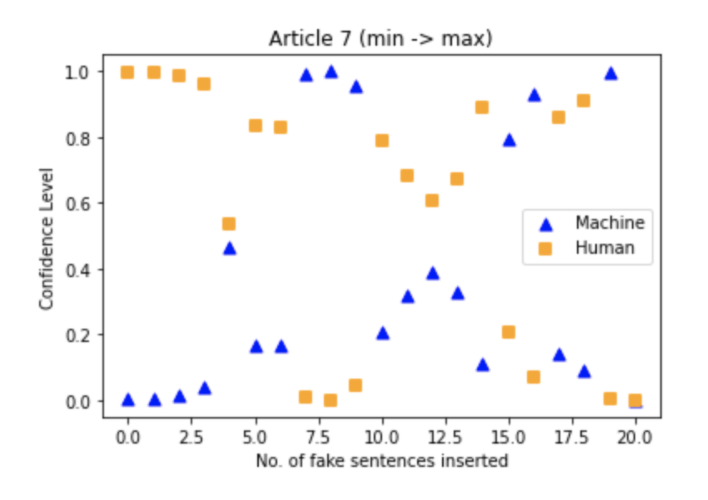}
     \end{subfigure}
     \hfill
     \begin{subfigure}[b]{0.4\textwidth}
         \centering
         \includegraphics[width=\textwidth]{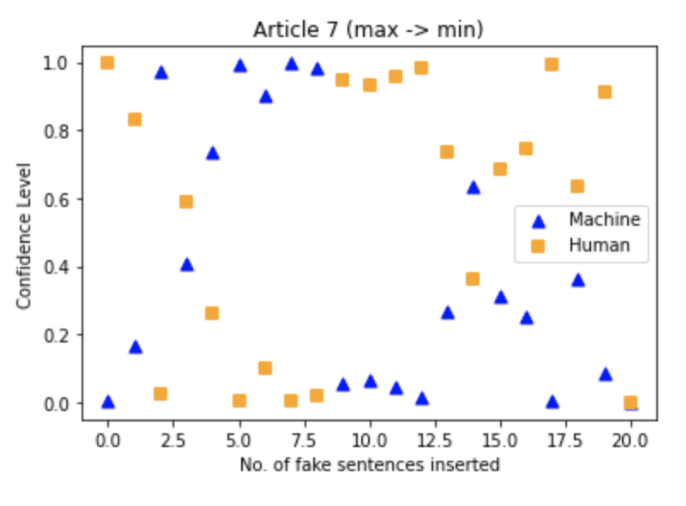}
     \end{subfigure}
    \caption{Effect of Subjectivity}
    \label{fig:subj}
\end{figure}

We generated the dataset for testing the effect of subjectivity through inserting sentences from the machine-written article (the fake sentences) into the human-written one. However, we insert them following a certain order. Before inserting, we sorted the fake sentences by their subjectivity values from the smallest to largest. We inserted the fake sentences following this order -- inserting the least subjective sentences -- to generate a series of articles consisting different proportions of fake sentences. We then do the same thing with the reversed order -- inserting the most subjective sentences first. Therefore, since the machine-generated articles tend to be more subjective, we would expect that if we insert the most subjective sentences first, the confidence level of the "human" class would start to drop sooner, compared to the articles where the least subjective sentences are inserted first. Results are shown in \ref{fig:subj}.

Taking article 7 as an example, as in \ref{fig:subj}, it appears that Grover has some threshold for the amount of subjectivity considered "acceptable" for classification purposes. Since most human-written articles are either entirely subjective (editorials, opinionated articles, etc.) or entirely objective (news reporting, etc.), we can see why Grover treats articles at either extremes as real. We see a significant drop in Grover's confidence score for an article being human-written after a certain number of subjective sentences are inserted. However, after some more insertions, Grover becomes less sure and we see the model jumping between human and machine-written classification.

\subsection{Word-Level Substitutions}
In order to assess the performance of Grover at the word level, we tested the effect of swapping words with their synonyms within articles. For each article (having length \textit{n} words), a word was chosen at random and was replaced with a synonym using the NLTK wordnet. The new article was then saved in a JSON file and the process was repeated again. This procedure produced a set of \textit{n} articles, where the first article had one synonym swapped, the second article had two synonyms, and so on. The set was run through the Grover discriminator to test the model's performance on each article.

\begin{figure}[htbp]
     \centering
     \begin{subfigure}[b]{0.4\textwidth}
         \centering
         \includegraphics[width=\textwidth]{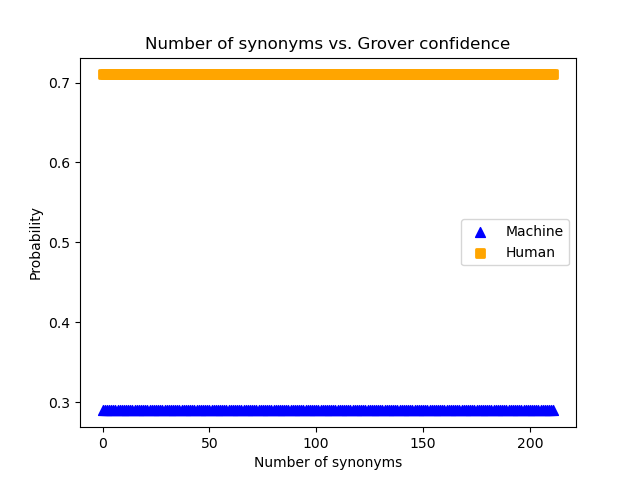}
         \caption{Human-Written Article}
         \label{fig:syn_human}
     \end{subfigure}
     \hfill
     \begin{subfigure}[b]{0.4\textwidth}
         \centering
         \includegraphics[width=\textwidth]{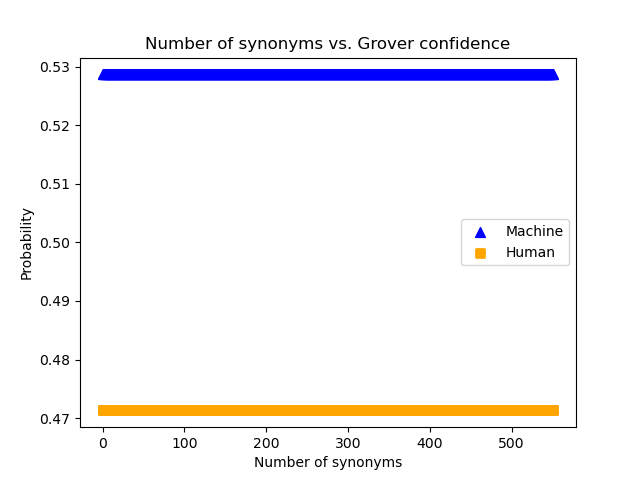}
         \caption{Machine-Written Article}
         \label{fig:syn_machine}
     \end{subfigure}
     \caption{Synonym Substitution}
\end{figure}

As seen in Figure \ref{fig:syn_human} and Figure \ref{fig:syn_machine}, swapping synonyms in this way had no effect on Grover's ability to classify articles. This finding was expected for the case of machine-written articles, as swapping synonyms does nothing to make an underlying article more "human." However, in the human-written case, the result was surprising. Similar to the sentence swap experiment in section \ref{sentence_swap}, we expected Grover's classification to shift toward 'Machine' as more synonyms were substituted in an article. 

The fact that the classification remained the same is likely due to two factors. First, only some of the words in each article had true synonyms. Proper nouns, for example, were unable to be substituted. Likewise, certain parts of speech had no synonyms. Articles such as \textit{a} and \textit{the}, and conjunctions like \textit{but} or \textit{and} are examples of such words. On average, we were only able to generate synonyms for fifty to sixty percent of the words in a given article. It is possible that swapping a higher number of words would lead to the expected effect. The second factor is related to how Grover perceives an article's "flow" when classifying. As discussed in sections \ref{sentence_swap} and \ref{sentence_insertion}, Grover is heavily influenced by changes at the discourse level. It could be that swapping synonyms in the above way doesn't impact the underlying structure of an article enough for Grover to perceive the changes as meaningful. While these two factors are likely, More research is needed in order to determine exactly why Grover behaves in this way.

\subsection{GPT-2 Substitutions}
The original Grover paper particularly emphasized the dangers of pre-existing language models like GPT-2 and BERT, which can also be used to generate fake news. Additionally, they noted that Grover performed extremely well on GPT-2 generated articles even though it had never seen GPT-2 generated articles during training, achieving 96.1\% accuracy \cite{Zellers2019-medium}. Given that Grover was not trained on GPT-2, and the testing done in the original experiment was on a relatively small set of data (500 generations), we felt that exploring perturbations using GPT-2 generations could yield promising results.
\subsubsection{Setup}
Perturbations using GPT-2 were all done using Deep AI's Text Generation API \cite{deepAI}. The generator is not the original GPT-2 model but rather a modified version of the original, and provides capabilities for word, sentence, and paragraph completion. For the purposes of our experiment, we refer to generations using this modified model as still being GPT-2 generations. We chose to explore sentence and paragraph completion, since they provide more context than single word generation and are more representative of an attack someone would want to perform to fool a reader.

For the experiment we used 8 original human-written articles, and replaced portions of the articles using GPT-2 generations, gradually increasing the percentage of the article that is GPT2-generated. First, the original article is loaded in and a sentence or paragraph is extracted. That text is then fed into the API and a GPT2-generated text is returned (usually multiple sentences or paragraphs). Next, the first sentence or paragraph is extracted from the generated text, and quotations/apostrophes replaced so it can be properly tokenized. Finally, the first sentence or paragraph occurring after the original piece of text is replaced with the generation, since the model works in a predictive fashion.
\subsubsection{Results}
\begin{figure}[htbp]
     \centering
     \begin{subfigure}[b]{0.4\textwidth}
         \centering
         \includegraphics[width=\textwidth]{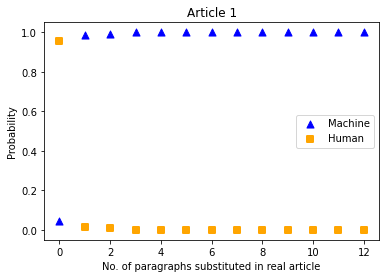}
         \caption{Article 1: Paragraph Substitution}
         \label{fig:par1}
     \end{subfigure}
     \hfill
     \begin{subfigure}[b]{0.4\textwidth}
         \centering
         \includegraphics[width=\textwidth]{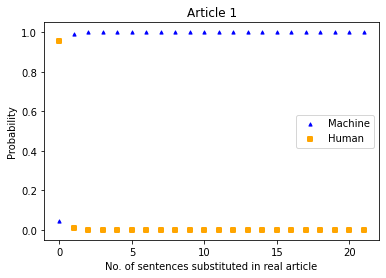}
         \caption{Article 1: Sentence Substitution}
         \label{fig:sent1}
     \end{subfigure}
     \hfill
     \begin{subfigure}[b]{0.4\textwidth}
         \centering
         \includegraphics[width=\textwidth]{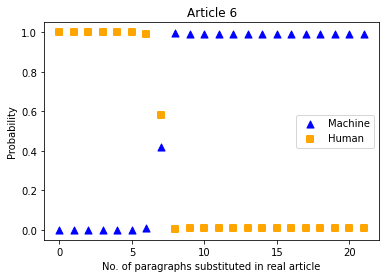}
         \caption{Article 6: Paragraph Substitution}
         \label{fig:par6}
     \end{subfigure}
    \hfill
    \begin{subfigure}[b]{0.4\textwidth}
         \centering
         \includegraphics[width=\textwidth]{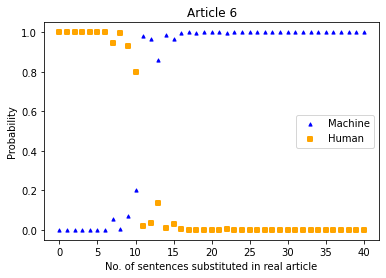}
         \caption{Article 6: Sentence Substitution}
         \label{fig:sent6}
     \end{subfigure}
        \caption{GPT-2 paragraph and sentence substitution classifications for articles 1 and 6}
\end{figure}

Overall, substituting GPT-2 generations into real articles showed varying but generally promising results. For both sentence and paragraph replacement, certain articles were classified as machine-generated very early, but most were not classified as such until a significant portion of the article was replaced. Additionally, while there did exist slight variations between sentence and paragraph replacement, the two produced very similar results and classification patterns, which reinforces both that the GPT-2 generator is good at both sentence and paragraph generation, and that similar context for the language model is provided by both sentences and paragraphs, and both are greater than that of word substitution. In Figures \ref{fig:par1} and \ref{fig:sent1} we see that both paragraph and sentence replacement for article 1 were detected immediately, representing good detection by Grover. Conversely, Figures \ref{fig:par6} and \ref{fig:sent6} show that a significant portion of article 6 had to be replaced before Grover consistently classified it as machine-written. All four figures demonstrate the similar classification patterns that were often seen between the two substitution types.

\begin{figure}[htbp]
     \centering
     \begin{subfigure}[b]{0.4\textwidth}
         \centering
         \includegraphics[width=\textwidth]{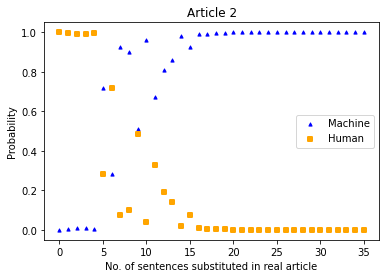}
         \caption{Article 2: Sentence Substitution}
         \label{fig:sent2}
     \end{subfigure}
     \hfill
     \begin{subfigure}[b]{0.4\textwidth}
         \centering
         \includegraphics[width=\textwidth]{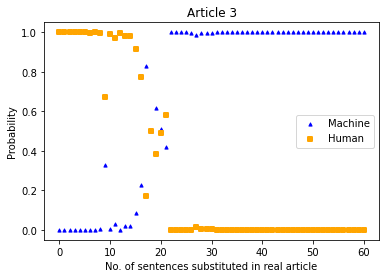}
         \caption{Article 3: Sentence Substitution}
         \label{fig:sent3}
     \end{subfigure}
        \caption{Individual sentence substitution strength anomalies for articles 2 and 3}
    \label{fig:sent23}
\end{figure}

We also observed an interesting phenomenon can be seen when replacing the sentences of article 2 and 3, as depicted in Figure \ref{fig:sent23}. At moments during the replacement, it seems as if Grover has "caught on" (classified as machine-written with high probability) that the article is partially machine written. Then, an effectively generated GPT-2 sentence causes the model to go back to classifying incorrectly, as is the case with the 6th and 21st substituted sentences for articles 2 and 3, respectively. This demonstrates that GPT-2 substitutions can have vastly different (and sometimes surprisingly effective) effects on Grover's ability to classify. This is likely because each substitution generation is based on a certain context; if Grover was trained on certain contexts and sequences of words, then it will perform better on them versus those that it hasn't seen before. This again reiterates the benefit of GPT2-based attacks, since Grover has little prior experience with them.

Ultimately though, our experiments show that despite certain variations, Grover is generally performs badly up to a certain portion of substitutions. To measure this, we define a value--undetected substitution percentage--which is the percentage of paragraphs or sentences substituted before Grover was able to consistently classify the modified article as machine-generated. As shown in Table \ref{table:usp}, the mean undetected substitution percentage was 24.4\% for paragraph substitution and 23.1\% for sentence substitution, with individual article values ranging from less than 10\% to over 35\% for both types. While our experiments were conducted on a small set of articles, these results indicate that attacks on small portions of the article are \textbf{unlikely} to be detected by Grover's classifier. In a real world situation, this means that an attacker could use GPT-2 to replace small but important pieces of an article such as the introduction, conclusion, or topic sentences in paragraphs to better fool humans without having to generate an entirely fake article.
\begin{table}[htbp]
    \centering
    \begin{tabular}{|c|c|c|c|c|c|c|c|c|c|}
    \hline
    Sub Type & A1 & A2 & A3 & A4 & A5 & A6 & A7 & A8 & Mean\\
    \hline
    Paragraph & 7.7\% & 23.8\% & 23.3\% & 33.3\% & 28.6\% & 36.4\% & 16.7\% & 25.0\% & 24.4\%\\
    Sentence & 4.5\% & 19.4\% & 36.1\% & 19.2\% & 23.8\% & 26.8\% & 19.4\% & 35.7\% & 23.1\%\\
    \hline
    \end{tabular}
    \centering
    \caption{Mean and individual article undetected substitution percentage}
    \label{table:usp}
\end{table}
\section{Informed Perturbations}
For the informed perturbations, we would like to utilize the embedding table of the Grover model. Similar to generating adversarial examples for images, we could apply the Fast Gradient Signed Method \cite{goodfellow2014explaining} to generate the adversarial embedding table. We first input the sentences to be perturbed into the classifier and get the loss. We now have the gradient of loss w.r.t. the embedding table. Then we choose different steps ($\epsilon = 0.001, 0.1, 1, 5$) times the gradient and add that to the embedding table. For all the tokens in the input sentences, there is gradient information for all those tokens. We then try to find a different token whose original embedding is similar to the adversarial embedding of the token. We choose the cosine similarity of the normalized embeddings and during the experiments, we found that a threshold of 0.18 is great to find similar embeddings while only changing one or two words in the original article. After we find the token pair to replace, we encode the original article to get the tokens, use the corresponding token to replace the attacked token, and decode it to a new article. One benefit is that since we are changing the same tokens, the probability distribution is unchanged. Therefore, the perplexity remains the same after perturbations. The token pair finding algorithm is shown in Algorithm \ref{alg:embed}.

\begin{algorithm}
	\caption{Find token pairs to replace} 
	\label{alg:embed}
	\begin{algorithmic}[1]
		\State E = embedding\_table
        \State G = embedding\_table\_gradient
        \State replace\_dict = \{\}
        \State adversarial\_embedding\_table = embedding\_table + step * embedding\_gradient
        \For {$each\ token\ used$}
            \State $A_T$ = token's adversarial embedding
            \State Find embedding $T$ in embedding table which has the maximum similarity with $A_T$
            \State $a$ is the token corresponding to embedding $T$
            \If {similarity($T, A_T$) $>$ threshold}
            \State replace\_dict.add((token, a))  // replace token with a     
            \EndIf
        \EndFor
	\end{algorithmic} 
\end{algorithm}

Empirically, for the articles we worked on, we successfully perturbed the classification result of the discriminator. Two examples are shown in Figure \ref{fig:r2f} and \ref{fig:f2r}. The first example fools the discriminator to tell the real news fake, while the second example fools the discriminator to tell the fake news real. Although it is more difficult to fool the discriminator to tell a machine-written article to be a human-written one, our method also succeeds in this direction. A simple guess is that the discriminator may believe that there are more complex words in human-written articles. 

In order to get the gradient information, we rewrite the relevant code of the Grover model. Specifically, we change the version of TensorFlow from 1.13.1 to 2.5. Originally, their input function and model function are implemented as function closure, which are then fed into the estimator. In order to get the gradient information in the model, we change the graph-based structure to eager mode. What's more, in order to load the original model weights, we deal with the problem of compatibility so we can load the original weight into our new model. We hope this can be beneficial for the research community if they would like to do further work on Grover.

\noindent
\begin{figure}[htbp]
\fbox{%
    \parbox{.46\textwidth}{%
  New electronic display technology can automatically correct for vision defects without glasses or contact lenses, according to {\color{magenta} researchers} at the MIT Media Lab and University of California at Berkeley.
The technology has applications for everything from e-readers and tablets to smartphones and GPS displays. By building vision correction into electronic displays {\color{magenta} researchers} are hoping to improve conditions in emerging markets where glasses and prescription lenses don't come easy.
Gordon Wetzstein, a researcher at MIT's Media Lab, penned a research paper describing the technology. The paper will be presented at Siggraph, a graphics conference, later this month. Ramesh Raskar, director of the Media Lab's Camera Culture group, and Berkeley’s Fu-Chung Huang and Brian Barsky are also listed on the paper.
The vision correction screens are a spin on the glasses-free 3D technology also developed at MIT. MIT's 3D screen projects different images to the left and right eye. The vision-correction version slightly different images to parts of the viewer's pupil.
    }%
}
\hfill
\fbox{%
    \parbox{.46\textwidth}{%
  New electronic display technology can automatically correct for vision defects without glasses or contact lenses, according to {\color{magenta} advoc} at the MIT Media Lab and University of California at Berkeley.
The technology has applications for everything from e-readers and tablets to smartphones and GPS displays. By building vision correction into electronic displays {\color{magenta} advoc} are hoping to improve conditions in emerging markets where glasses and prescription lenses don't come easy.
Gordon Wetzstein, a researcher at MIT's Media Lab, penned a research paper describing the technology. The paper will be presented at Siggraph, a graphics conference, later this month. Ramesh Raskar, director of the Media Lab's Camera Culture group, and Berkeley’s Fu-Chung Huang and Brian Barsky are also listed on the paper.
The vision correction screens are a spin on the glasses-free 3D technology also developed at MIT. MIT's 3D screen projects different images to the left and right eye. The vision-correction version slightly different images to parts of the viewer's pupil.
}%
}

\caption{The real news (left) is perturbed to become the fake news (right). The original classification result is that the news is human-written with a probability of 83\%. We replace 'researchers' with 'advoc'. The classification result after perturbations is that the news is machine-written with a probability of 62\%. }
\label{fig:r2f}
\end{figure}

\noindent
\begin{figure}[htbp]
\fbox{%
    \parbox{.46\textwidth}{%
  HAVANA, Cuba — The Obama administration has approved the {\color{magenta} first} U.S. factory in Cuba in more than 50 years, allowing a two-man company from Alabama to build a plant assembling as many as 1,000 small tractors a year for sale to private farmers in Cuba.
The Treasury Department last week notified partners Horace Clemmons and Saul Berenthal that they can legally build tractors and other heavy equipment in a special economic zone started by the Cuban government to attract foreign investment.
Cuban officials already have publicly and enthusiastically endorsed the project. The partners said they expect to be building tractors in Cuba by the {\color{magenta} first} quarter of 2017.
"It's our belief that in the long run we both win if we do things that are beneficial to both countries," said Clemmons.
The $5 million to $10 million plant would be the {\color{magenta} first} significant U.S. business investment on Cuban soil since Fidel Castro took power in 1959 and nationalized billions of dollars of U.S. corporate and private property. 
    }%
}
\hfill
\fbox{%
    \parbox{.46\textwidth}{%
  HAVANA, Cuba — The Obama administration has approved the {\color{magenta} Winchester} U.S. factory in Cuba in more than 50 years, allowing a two-man company from Alabama to build a plant assembling as many as 1,000 small tractors a year for sale to private farmers in Cuba.
The Treasury Department last week notified partners Horace Clemmons and Saul Berenthal that they can legally build tractors and other heavy equipment in a special economic zone started by the Cuban government to attract foreign investment.
Cuban officials already have publicly and enthusiastically endorsed the project. The partners said they expect to be building tractors in Cuba by the {\color{magenta} Winchester} quarter of 2017.
"It's our belief that in the long run we both win if we do things that are beneficial to both countries," said Clemmons.
The $5 million to $10 million plant would be the {\color{magenta} Winchester} significant U.S. business investment on Cuban soil since Fidel Castro took power in 1959 and nationalized billions of dollars of U.S. corporate and private property. 
}%
}

\caption{The fake news (left) is perturbed to become the real news (right). The original classification result is that the news is machine-written with a probability of 81\%. We replace 'first' with 'Winchester'. The classification result after perturbations is that the news is human-written with a probability of 66\%. }
\label{fig:f2r}
\end{figure}

\clearpage

\section{Conclusion and Further Work}
Grover is an incredibly robust model that has proven to be great at both generating an detecting neural fake news; however, our experiments demonstrate that certain targeted attacks against Grover do indeed reduce its performance and leave it vulnerable to similar attacks in the real world. Out of the approaches we tried, we found that Grover-generated substitution, GPT2-generated substitution, length changing, and perturbations informed with the model's embedding table were all effective to some degree in fooling Grover's detection. In addition, our research demonstrated that synonym substitution, fake-fake substitution, insertion instead of substitution, and changing the position of substitution, all had little to no effect on Grover's detection abilities. Another takeaway from our research project is the difficulty in fooling Grover to think a fake article is real. While it is quite easy to fool Grover to classify a real article as fake, only adversarial embedding could fool Grover to classify a fake article as real. We rewrote the code in TensorFlow 2 and solved the issue of checkpoint compatibility, which is beneficial for further research on Grover.

Future work first entails doing more thorough testing through a larger variety of attacks and with more data. For most of the experiments we conducted, we were working with only a few articles, so performing the same tests on a much larger amount of data would reinforce or potentially invalidate some of the conclusions we came to in our small scale experiments. Additionally, we could try different adversarial approaches such as perturbations using BERT or the new GPT-3 which is significantly more robust than GPT-2. 

\clearpage

\bibliographystyle{plain}
\bibliography{bibliography}

\begin{thebibliography}{10}

\bibitem{Alcott2017}
Hunt Allcott and Matthew Gentzkow.
\newblock Social media and fake news in the 2016 election.
\newblock {\em Journal of Economic Perspectives}, 31,2:211--36, 2017.

\bibitem{Boberg2019}
Thorsten Boberg and Tim Schatto-Eckrodt.
\newblock Fake news.
\newblock {\em The International Encyclopedia of Journalism Studies}, 2019.

\bibitem{deepAI}
DeepAI.
\newblock {GPT2-Based Text Generator}.
\newblock https://deepai.org/machine-learning-model/text-generator.

\bibitem{Devlin2018}
Jacob Devlin, Ming-Wei Chang, Kenton Lee, and Kristina Toutanova.
\newblock Bert: Pre-training of deep bidirectional transformers for language
  understanding.
\newblock {\em arXiv preprint arXiv:1810.04805}, 2018.

\bibitem{Faris2017}
Robert Faris, Hal Roberts, Nikki~Bourassa Bruce~Etling, Ethan Zuckerman, and
  Yochai Benkler.
\newblock Partisanship, propaganda, and disinformation: Online media and the
  2016 us presidential election.
\newblock {\em Berkman Klein Center Research Publication}, 2017.

\bibitem{goodfellow2014explaining}
Ian~J Goodfellow, Jonathon Shlens, and Christian Szegedy.
\newblock Explaining and harnessing adversarial examples.
\newblock {\em arXiv preprint arXiv:1412.6572}, 2014.

\bibitem{Joulin2017}
Armand Joulin, Edouard Grave, Piotr Bojanowski, , and Tomas Mikolov.
\newblock Bag of tricks for efficient text classification.
\newblock In {\em Proceedings of the 15th Conference of the European Chapter of
  the Association for Computational Linguistics: Volume 2, Short Papers},
  volume~2, pages 427--431. Association for Computational Linguistics, 2017.

\bibitem{Lazer2018}
M.~J. et~Al Lazer, David.
\newblock The science of fake news.
\newblock {\em Science}, 359,6380:1094--1096, 2018.

\bibitem{Radford2019}
Alec Radford, Jeffrey Wu, Rewon Child, David Luan, Dario Amodei, and Ilya
  Sutskever.
\newblock Bert: Pre-training of deep bidirectional transformers for language
  understanding.
\newblock Technical report, OpenAI, 2019.

\bibitem{Reis2019}
Julio et~Al Reis.
\newblock Supervised learning for fake news detection.
\newblock {\em IEEE Intelligent Systems}, 34,2:76--81, 2019.

\bibitem{Shu2017}
Kai Shu, Amy Sliva, Suhang Wang, Jiliang Tang, and Huan Liu.
\newblock Fake news detection on social media: A data mining perspective.
\newblock {\em ACM SIGKDD Explorations Newsletter}, 19,1:22--36, 2018.

\bibitem{Vosoughi2018}
Soroush Vosoughi, Deb Roy, and Sinan Aral.
\newblock The spread of true and false news online.
\newblock {\em Science}, 359,6380:1146--1151, 2018.

\bibitem{Wardle2018}
Claire Wardle and Hossein Derakhshan.
\newblock Thinking about ‘information disorder’: Formats of misinformation,
  disinformation, and mal-information.
\newblock In {\em Journalism, ‘Fake News’ \& Disinformation}. UNESCO, 2018.

\bibitem{Weir2009}
William Weir.
\newblock {\em History's Greatest Lies}.
\newblock Fair Winds Press, 2009.

\bibitem{Zellers2019-medium}
Rowan Zellers, Ari Holtzman, Hannah Rashkin, Yonatan Bisk, Ali Farhadi,
  Franziska Roesner, and Yejin Choi.
\newblock Counteracting neural disinformation with grover.
\newblock {\em Medium}, 2019.

\bibitem{Zellers2019-arXiv}
Rowan Zellers, Ari Holtzman, Hannah Rashkin, Yonatan Bisk, Ali Farhadi,
  Franziska Roesner, and Yejin Choi.
\newblock Defending against neural fake news.
\newblock {\em arXiv preprint arXiv:1905.12616}, 2019.

\bibitem{Zhou2019}
Xinyi et~Al Zhou.
\newblock Fake news: Fundamental theories, detection strategies, and
  challenges.
\newblock In {\em WSDM '19: Proceedings of the Twelfth ACM International
  Conference on Web Search and Data Mining}, pages 836--37, 2019.

\end{thebibliography}

\end{document}